\documentclass[conference]{IEEEtran}
\IEEEoverridecommandlockouts
\usepackage{cite}
\usepackage{amsmath,amssymb,amsfonts}
\usepackage{algorithmic}
\usepackage{graphicx}
\usepackage{textcomp}
\usepackage{cuted}
\usepackage{capt-of}
\usepackage{float}
\usepackage{booktabs}
\usepackage{threeparttable}
\usepackage{mathtools}
\usepackage[mathscr]{eucal}
\usepackage[table,xcdraw]{xcolor}
\usepackage{colortbl}
\usepackage{colortbl}
\usepackage{balance}
\usepackage{color}
\usepackage{multirow}
\usepackage{multicol}
\usepackage{marvosym}
\usepackage{algorithmic}
\usepackage{indentfirst}
\usepackage{makecell}
\usepackage{caption}
\usepackage{threeparttable}
\def\BibTeX{{\rm B\kern-.05em{\sc i\kern-.025em b}\kern-.08em
    T\kern-.1667em\lower.7ex\hbox{E}\kern-.125emX}}
\begin{document}

\title{ConsistencyPlanner: Real-time Planning with Fast-Sampling Consistency Models \\
}

\author{
Qichao Zhang$^{1,2}$,
Xing Fang$^{1}$,  
Jiaqi Fang$^{3}$,
Zhenwen Cai$^{3}$,
Jie Ling$^{3}$,
Qiankun Yu$^{3}$,
and Dongbin Zhao$^{1,2}$\textsuperscript{\Letter} \\
\textsuperscript{1}State Key Laboratory of Multimodal Artificial Intelligence Systems, \\Institute of Automation, Chinese Academy of Sciences, Beijing 100190, China \\
\textsuperscript{2}School of Artificial Intelligence, University of Chinese Academy of Sciences, Beijing 100049, China\\
\textsuperscript{3}Guangzhou Zaofu Intelligent Technology Co., Ltd.\\
\textsuperscript{\Letter} Corresponding Author
}

\maketitle

\begin{abstract}

Closed-loop planning in complex, real-world driving scenarios presents a critical challenge for autonomous driving systems. While traditional rule-based methods are interpretable, their predefined heuristics lack the adaptability for dynamic traffic environments. 
Learning-based approaches have shown considerable promise. 
Conversely, learning-based approaches, despite their promise, struggle to balance the modeling diverse and multimodal driving behaviors and real-time planning, often leading to indecisive or unsafe actions.
To address this limitation, we propose \textit{ConsistencyPlanner}, a real-time planning framework with fast-sampling consistency models. Our approach is built upon two key technical contributions.
\textit{Efficient Multimodal Sampling:} We employ fast-sampling consistency models to generate a diverse set of plausible future trajectories. This enables efficient, real-time exploration of multimodal actions, overcoming the computational bottlenecks of previous iterative generative methods.
\textit{Heterogeneous Feature Fusion:} We introduce an attention-enhanced decoder that dynamically integrates heterogeneous input features—including scene feature and action token—into a cohesive representation for robust planning.
 Extensive evaluation in the Waymax simulator demonstrates superior performance in safety metrics compared to existing methods, with particularly strong results in challenging dynamic scenarios.
\end{abstract}


\section{Introduction}
The pursuit of safe and efficient autonomous driving has precipitated a paradigm shift from open-loop prediction~\cite{nayakanti2023_irca_wayformer} to closed-loop planning~\cite{dauner23a_corl_pdm, Zhengyupeng}. 
This transition has been further accelerated by the emergence of closed-loop planning benchmarks tailored for real-world driving scenarios, such as Waymax~\cite{Gu2023_nips_waymax}, which have refocused the field from isolated prediction tasks to the more complex, interactive challenges inherent in closed-loop control.
Within this context, current planning approaches can be broadly categorized into rule-based and learning-based methods. Rule-based systems rely on carefully engineered heuristics to govern vehicle behavior, offering strong controllability and interpretability in decision-making, as exemplified by platforms like Apollo~\cite{fan2018_baidu_applo}. However, their dependence on predefined logic often hampers adaptability in novel or dynamically complex urban environments, where unexpected situations may fall outside the scope of handcrafted rules. In contrast, learning-based methods, including models such as GameFormer~\cite{huang2023_iccv_gameformer}, learn driving policies directly from large-scale human demonstrations. By leveraging increasingly extensive datasets and powerful neural architectures like Transformers~\cite{vaswani2017_nips_transformer}, these methods demonstrate enhanced generalization to diverse and unstructured real-world conditions.

While learning-based approaches have demonstrated considerable promise in autonomous driving ~\cite{yang2026worldrft, zheng2025world4drive}, they still face critical challenges in real-world deployment. A key limitation is their inability to  capture the inherent multimodality and uncertainty of complex driving scenarios~\cite{nayakanti2023_irca_wayformer,li2023planning}. Although behavior cloning techniques~\cite{yang2025uncad} can approximate expert demonstrations, they offer no theoretical guarantee of accurately representing the diversity of human decision-making, even when supported by large-scale transformer architectures~\cite{vaswani2017_nips_transformer}. 
Recent advances have explored generative models to improve multimodal expressiveness. For example, LatentDriver~\cite{xiao2024_arxiv_latentdriver} explores the feasibility of deriving decisions from an autoregressive world model through the formulation of multiple probabilistic hypotheses on Waymax. Diffusion Planner~\cite{zheng2025_iclr_diffusion_planner} employs the diffusion model as the policy and achieve superior performances on the Nuplan Benchmark~\cite{caesar2021nuplan}. 
While these models demonstrate stronger capability in generating diverse trajectories, their substantial computational overhead during inference poses a serious barrier to real-time deployment. The resulting latency is incompatible with the low-latency requirements of practical autonomous driving systems. 

Building upon recent advances in generative modeling, we identify consistency models~\cite{song23a_icml_cm} as a particularly promising solution to the dual challenges of multi-modal behavior representation and real-time inference in autonomous driving. These models combine the distribution modeling capabilities of diffusion processes~\cite{chen2023_aamas_cpql,prasad2024_consistency_policy} with efficient single-step sampling, making them uniquely suited for capturing the diverse decision space of human drivers while meeting stringent latency requirements. We present \textit{ConsistencyPlanner}, the framework to successfully adapt fast-sampling consistency models for closed-loop autonomous planning with real-world driving dataset. Specifically, we propose a new network architecture built upon the diffusion transformer~\cite{Peebles2023_iccv_dit} to fuse different input features. Evaluation results on the large-scale real-world autonomous planning benchmark Waymax~\cite{Gu2023_nips_waymax} demonstrate that \textit{ConsistencyPlanner} achieves superior closed-loop performance among learning-based baselines.

In this paper, we demonstrate that, through appropriate architectural design, the potential of consistency models can be fully leveraged to enhance closed-loop planning performance in autonomous driving. The main contributions of this work are summarized as follows:  

\begin{itemize}  
    \item To balance multimodal planning with real-time inference, we integrate consistency models into our closed-loop planning framework. Driven by real-world scenarios, these models provide fast sampling capabilities that accurately capture the multimodal characteristic of driving behaviors. 
    \item We incorporate an attention mechanism into the consistency model decoder to achieve thorough fusion of input conditions and action information.  
    \item Experimental results on the Waymax simulator validate the effectiveness of our method and demonstrate superior performance, particularly on safety metrics. 
\end{itemize}

\section{Related Work}

\subsection{Diffusion Models and Consistency Models}

Diffusion models have emerged as the predominant paradigm for generative modeling, which operate through an iterative denoising process that progressively refines Gaussian noise into high-quality samples. The foundational DDPM framework~\cite{ho2020_nips_ddpm} establish this paradigm, with subsequent improvements in sampling efficiency~\cite{song2021scorebased} and quality control via classifier-free guidance~\cite{ho2022classifier_free}.
The field has undergone a paradigm shift with the advent of consistency models~\cite{song23a_icml_cm}. These models distill the iterative diffusion process into a single generation step via learned consistency mappings, thereby preserving the sample quality of traditional diffusion while enabling orders-of-magnitude faster inference. This breakthrough is particularly transformative for real-time applications. Subsequent advancements in training methodologies~\cite{song2024_iclr_improved_cm} have further solidified their efficacy, leading to successful deployment in diverse domains such as video~\cite{videolcm}, 3D motion generation~\cite{motionlcm}, and robotic tasks~\cite{prasad2024_consistency_policy,li2024_nips_cp3er}, where computational efficiency is paramount.

\subsection{Diffusion Model for Autonomous Driving}

Diffusion models~\cite{ho2020_nips_ddpm} have emerged as a powerful class of generative frameworks that synthesize data through a learned denoising process. Their ability to model complex, multimodal data distributions has made them particularly attractive for robot tasks~\cite{yaochyeng2024_TAI}. In recent years, these models have gained increasing traction in the field of autonomous driving~\cite{zheng2024planagent}, where tasks such as trajectory forecasting and traffic scene generation inherently involve uncertainty and multimodality. For instance, Jiang et al.~\cite{jiang23_cvpr_motiondiffuser} apply diffusion-based techniques to forecast future trajectories of surrounding agents, including vehicles and pedestrians, effectively capturing the diverse range of plausible future behaviors. Similarly, works such as~\cite{zhonge23_icra_ctg,chitta24_eccv_sledge} extend diffusion models to traffic simulation, demonstrating their capability to produce realistic and controllable environments for downstream testing and training. Beyond perception and simulation, diffusion models are also being explored for decision-making and planning. Yang et al.~\cite{yang2024_cvpr_diffusion_es} propose a diffusion-based framework that generates a set of candidate trajectories, which are subsequently scored by pre-trained evaluators~\cite{dauner23a_corl_pdm} to select the final motion plan. Notably, Zheng et al.~\cite{zheng2025_iclr_diffusion_planner} introduce a fully closed-loop planning system based on diffusion models, showing competitive or superior performance compared to traditional rule-based approaches, even without post refinement. Despite their promising capabilities, a major limitation of diffusion models is their reliance on iterative sampling procedures involving hundreds of denoising steps. This requirement significantly limits their suitability for real-time deployment in safety-critical systems like autonomous vehicles, where low-latency planning is essential.

\section{Method}

\subsection{Preliminary}

Consistency models provide an efficient alternative to traditional diffusion models by learning a time-consistent mapping from noisy inputs to clean data. Unlike diffusion models that require iterative denoising, consistency models enable one-step generation by enforcing consistency across different time steps of the forward diffusion process. Consider the stochastic differential equation (SDE) that defines the forward process:
\begin{equation}
    dx^t = \mu(x^t, t)dt + \sigma(t)dw^t,
\end{equation}
where $x^0 \sim p_{data}(x)$ is a clean sample and $x^T \sim \mathcal{N} (\textbf{0}, T^{2} \textbf{I})$ denotes the final noisy data.

A consistency model learns a function $F_\theta(x^t, t)$ satisfying the \emph{consistency condition}:$F_\theta(x^t, t) \equiv x^0, \forall t \in [0, T],$ ensuring that all time steps along a diffusion path map to the same clean sample. To train $F_\theta$, the model minimizes the consistency loss:
\begin{equation}
    \mathcal{L}_{\text{CT}}(\theta) = \mathbb{E}_{x^0, t, t'}\left[ d(F_\theta(x^t, t),  F_\theta(x^{t'}, t'))\right],
\end{equation}
where $x^t$ and $x^{t'}$ are generated from the same $x^0$ at different time steps, $d$ represents the loss function. At inference, one-step sampling is performed by evaluating:$ \hat{x}^0 \sim F_\theta(x^T, T),$
offering a fast and practical generation strategy without iterative denoising.

\subsection{ConsistencyPlanner}

\subsubsection{Overview}

\begin{figure*}[t]
    \centering
    \includegraphics[width=0.97\textwidth]{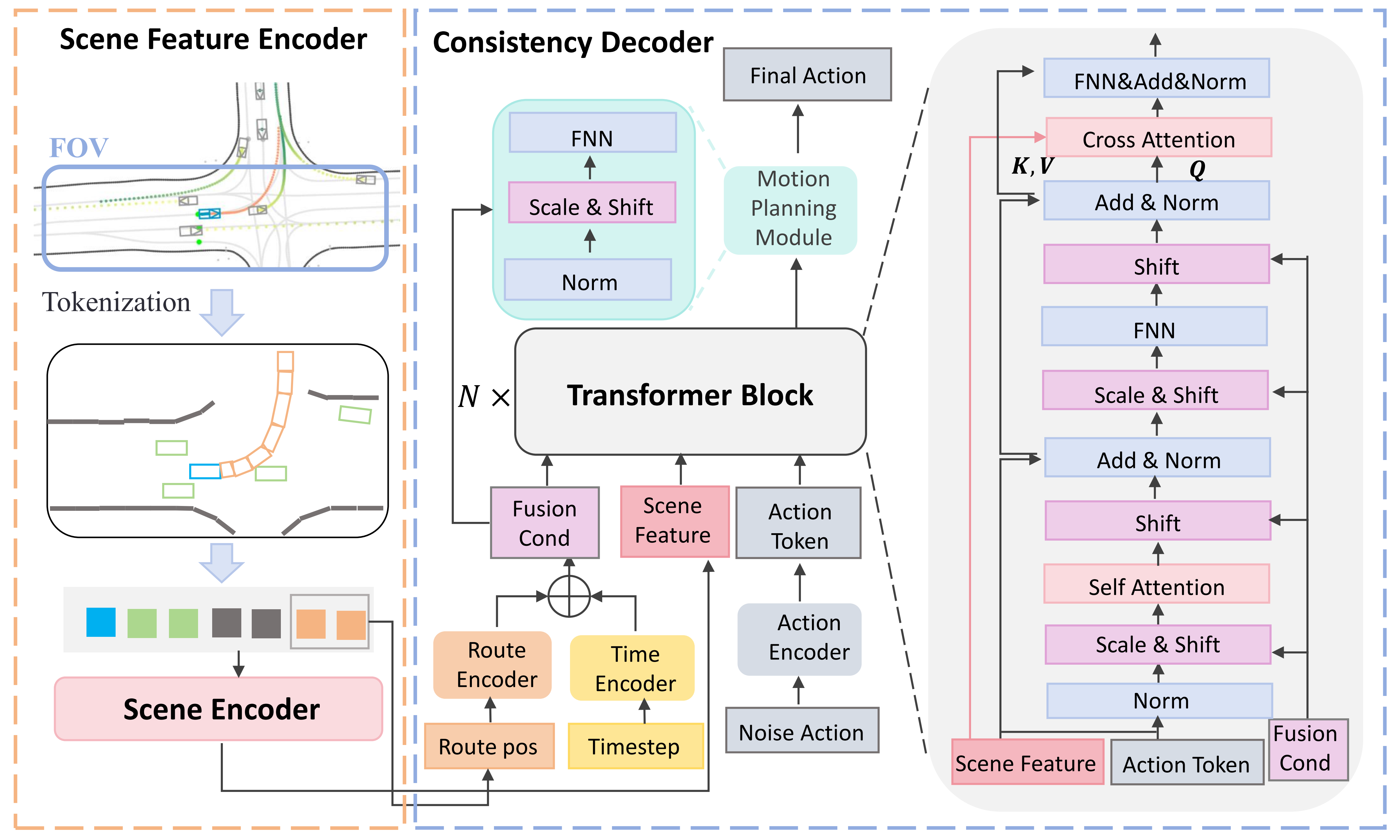}
    \caption{Framework of the ConsistencyPlanner method with the scene feature encoder and consistency decoder. We incorporate an transformer block into the consistency decoder to achieve thorough fusion of input conditions and action information.}
    \label{fig_framework}
\end{figure*}

As illustrated in Figure~\ref{fig_framework}, the ConsistencyPlanner framework consists of two key components: Scene feature encoder, and Consistency Decoder. For scene feature encoding, we first extract raw driving information from the WOMO dataset, including road features, route features, ego vehicle state, and surrounding vehicle states. However, such a comprehensive scene representation inherently contains redundancies. To mitigate this issue, we perform scene tokenization in the ego vehicle’s coordinate system, thereby obtaining a more condensed set of scene features. These tokenized features are subsequently processed by a BERT encoder~\cite{devlin2019_bert} to generate compact and unified representations for the decoder.

The consistency decoder leverages consistency models to enable multimodal representation learning. Drawing inspiration from Diffusion Planner~\cite{zheng2025_iclr_diffusion_planner}, we highlight the pivotal role of route features in autonomous driving policy formulation. Specifically, route features are integrated with consistency timesteps to construct fused conditional inputs for the decoder. This unified framework facilitates expert-level motion planning generation through iterative denoising.

\subsubsection{Scene Feature Encoder}

At each timestep, we perform tokenization of the raw observations within the ego vehicle's coordinate frame, considering a Field of View (FOV) with predefined dimensions $w_f \times h_f$ as specified in~\cite{xiao2024_arxiv_easychauffeur}. This processing yields two distinct element categories: static elements (road maps and navigation trajectories) and dynamic elements (vehicular agents). For static elements, we employ the Ramer-Douglas-Peucker algorithm~\cite{ramer1972,douglas1973} to approximate dense coordinate sequences as parameterized rectangular regions $[x, y, w, h, \psi, \text{id}]$, where road features $S_{rd} \in \mathbb{R}^{N_{rd} \times 6}$ use a fixed 0.5m width while route features $S_{rt} \in \mathbb{R}^{N_{rt} \times 6}$ adopt the ego vehicle's width. Dynamic elements consist of neighboring agents represented as $S_{\text{neighbor}} \in \mathbb{R}^{N_{\text{neighbor}} \times 6}$ (with velocity replacing the id field) and the ego vehicle state $S_{\text{ego}} \in \mathbb{R}^{1 \times 6}$. The complete observation representation $O \in \mathbb{R}^{(N_{rd}+N_{rt}+N_{\text{neighbor}}+1) \times 6}$ integrates all features within the perceptual bounds, augmented with type indicators for distinction.

To integrate object-centered vectorized representations from the scene, we employ a BERT-based encoder~\cite{devlin2019_bert}. The encoder processes the observation tensor $O_t$ along with its associated class tokens, generating a compact scene embedding $\text{scene}_{\text{emb}} \in \mathbb{R}^d$ that encapsulates both the static and dynamic state representations.

\subsubsection{Consistency Decoder}

Route information plays a crucial role in guiding the driving policy. Following the EasyChauffeur~\cite{xiao2024_arxiv_easychauffeur}, we extract route features $S_{\text{route}} \in \mathbb{R}^{20 \times D_{\text{pos}}}$ by downsampling the logged trajectories, where $D_{\text{pos}}$ encodes the $(x, y, \psi)$ coordinates. These features are distinct from the encoded representation $S_{rt}$. The route features are further processed using stacked MLP-Mixer layers \cite{Tolstikhin2021_nips_mlp_mixer}, which perform bidirectional mixing across both the vector and feature dimensions:
\begin{equation}
S_{\text{route}} = S_{\text{route}} + \text{MLP}(S_{\text{route}}^T)^T, \quad S_{\text{route}} = S_{\text{route}} + \text{MLP}(S_{\text{route}})
\end{equation}
The consistency time step encoding module generates $t_{\text{emb}}$, and the route information encoding module produces $S_{\text{route}}$. These features, now in the same latent space dimension, are combined through addition to form the fused condition $\text{cond} = t_{\text{emb}} + S_{\text{route}}$. For the noisy action input $a_t$, it is encoded by an action information encoding module to obtain the latent action tokens $a_{\text{emb}}$.  Finally, the scene encoding features $\text{scene}_{\text{emb}}$, obtained from the scene encoder, are incorporated. All components are integrated into the Transformer block for different feature processing. We employ Adaptive Layer Normalization with Zero Initialization (AdaLN-Zero)~\cite{Peebles2023_iccv_dit} to achieve conditional feature fusion. 
The conditioned action embedding $a_{\text{emb}}$ is then combined with scene features through multi-head cross-attention:
\begin{equation}
    a_\text{emb} = \text{CrossAttn}(\text{query} = a_\text{emb}, \text{key} = \text{value} = \text{scene}_{\text{emb}})
\end{equation}
The final executable action is generated through conditioned feature projection, where the fusion condition modulates both scaling and shifting operations:
\begin{equation}
    \beta, \gamma = \text{MLP}(\text{cond}) \quad a_\text{final} = \text{FFN}(\beta \text{LayerNorm}(a_\text{emb}) + \gamma)
\end{equation}



\subsubsection{Training Loss}
We present a novel integration of consistency models~\cite{song23a_icml_cm} into closed-loop autonomous driving systems. The proposed consistency decoder architecture is formally defined as:
\begin{equation}
\label{equ_pi_w}
    \pi_{\theta}(a|s) \triangleq f_{\theta}(a^{t}, t, s)
     = c_{skip}(t)a^{t} + c_{out}(t)F_{\theta}(a^{t},t|s)
\end{equation}
where $t$ denotes the discrete timestep related to ordinary differential equations, $a^{t} \sim \mathcal{N}(\textbf{0}, t^2\textbf{I})$ represents the noisy action sampled from a Gaussian distribution with zero mean and covariance matrix $t^2\textbf{I}$, and the corresponding state $s$ in this chapter includes: $S_{\text{route}}$ and $\text{scene}_{\text{emb}}$. $F_{\theta}(a^{t},t|s)$ is the neural network model to be trained, which outputs a conditional action with the same dimensionality as $a^{t}$ based on the current input state information $s$. $c_{skip}(t)$ and $c_{out}(t)$ are defined differentiable functions.

The specific loss function for the consistency decoder is formulated as:
\begin{equation}
    L(\theta) = \mathbb{E}_{(s,a) \sim \mathcal{D}, z \sim \mathcal{N} (\textbf{0}, \textbf{I})} \left[ d(f_{\theta}(a + t_{m+1}z, t_{m+1} | s), a) \right]
\end{equation}
where the loss function $d$ employs the $L_1$ norm, and $t_{m+1}$ is obtained by discretizing the time interval $[\epsilon, T]$ using the Karra method~\cite{karras}. Here, $\epsilon$ is a small constant close to zero to avoid numerical instability, and $T$ represents the total timesteps of the consistency models. Finally, we use $a_\text{final} = \hat{a}_{\epsilon} \sim \pi_{\theta}(a|s)$ as the executed action.


\section{Experimental Results}
\label{sec:result}

\subsection{Experimental Dataset}

All experiments are carried out using Waymax~\cite{Gu2023_nips_waymax}, a recently introduced simulation platform based on the Waymo Open Motion Dataset (WOMD v1.1.0)~\cite{Ettinger2021_iccv_waymo}. Each scenario spans 8 seconds and is sampled at 10 Hz, resulting in a total of 80 frames per scenario. Agent control also operates at 10 Hz, supporting up to 128 agents simultaneously within a single scene. The training dataset consists of 487,002 scenarios, while 44,096 scenarios are reserved for validation. Simulations proceed for the full 8-second duration without early termination, ensuring consistent evaluation across all runs.

\subsection{ Evaluation Metrics}

All experiments are conducted under a closed-loop evaluation protocol, following the setup proposed in~\cite{Gu2023_nips_waymax}. The evaluation metrics include \textbf{Off-road Rate (OR)} and \textbf{Collision Rate (CR)}, both adopted directly from the original definitions in~\cite{Gu2023_nips_waymax}. The \textbf{Progress Ratio (PR)} is measured in a manner consistent with prior work; however, in this study, the maximum value is capped at 100\% due to the unavailability of future drivable area annotations in the WOMD dataset.

\subsection{Implementation Details}
For each frame of the dynamic driving scenario, our method discretizes the road information into $N_{rd}=40$ road feature rectangles and extracts $N_{rt}=20$ route information rectangles. The system considers $N_{vehicle}=128$ dynamic entities comprising the autonomous vehicle and its neighboring dynamic obstacles, resulting in $N_{\text{neighbor}}=127$ interaction targets. 

 In the network architecture design, the consistency model decoder employs $N_{\text{block}}=3$ stacked layers with $h=4$ attention heads, and the latent feature dimension is set to $D=256$. For the time step encoding module, we use a frequency embedding space of dimension $d_\text{freq}=256$ with a maximum time period $T_{\text{max}} = 10000$. The navigation information encoding module processes vector-dimensional features with 32 encoding dimensions and route features with 64 dimensions.

 All training procedures were conducted on 8 NVIDIA V100 GPUs using the Adam optimizer with the following hyperparameters: 20 training epochs, initial learning rate of $2\times10^{-4}$ with OneCycleLR scheduling policy.  During training, the sub-sequence \(\{t_m | m \in [T]\}\) differs from inference and follows the Karras boundary schedule~\cite{karras}:  
\[
t_m = \left(\epsilon^{1/\rho} + \frac{m-1}{T-1}\left(t_T^{1/\rho} - \epsilon^{1/\rho}\right)\right)^\rho,
\]  
where \(t_T\) denotes the terminal timestep, \(\epsilon\) is a small constant for numerical stability, and \(\rho\) controls the curvature of the schedule. The consistency model operates with $t_T=80$ total time steps, $\epsilon=0.002$ as the minimum noise level, $T=40$ and $\rho=7$.

\subsection{Performance Comparison}

\begin{table*}[htbp]
    \centering
    \caption{Comparison with state-of-the-art methods. Non-ego agents are controlled by IDM~\cite{treiber2000_idm}. `LT' and `DF' under \textbf{Route} indicate Logged Trajectory and Drivable Futures (unavailable), respectively.}
    \label{tab:waymax_comparision}
    \resizebox{\textwidth}{!}{
    \begin{tabular}{lccc|cc|c}
        \toprule
        \textbf{Methods} &\textbf{Venue} & \textbf{Route} & \textbf{Action Space} & \textbf{OR}$\downarrow$ & \textbf{CR}$\downarrow$ & \textbf{PR$^*$}$\uparrow$ \\
        \midrule
        Wayformer-BC~\cite{Gu2023_nips_waymax} & \multirow{3}{*}{NeurIPS 2023}  & LT+DF    & \multirow{2}{*}{Bicycle(Discrete)} & \textbf{1.11} & \textbf{4.59} & \textcolor{gray}{129.84} \\
        DQN~\cite{Gu2023_nips_waymax}   & & LT+DF   &                             & 3.74 & 6.50 & \textcolor{gray}{177.91} \\
        Waymax-BC~\cite{Gu2023_nips_waymax}      &   & LT+DF & \multirow{2}{*}{Bicycle}   & 13.59$\pm$12.71 & 11.20$\pm$5.34 & \textcolor{gray}{137.11$\pm$33.78} \\
        EasyChauffeur-PPO~\cite{xiao2024_arxiv_easychauffeur} &{Arxiv 2024} & LT    &                            & 3.95 & 4.72 & 98.26 \\
        \midrule
        Wayformer~\cite{nayakanti2023_irca_wayformer}   &{ICRA 2023}        & LT+DF & \multirow{5}{*}{Waypoints} & 7.89 & 10.68 & \textcolor{gray}{123.58} \\
        Waymax-BC~\cite{Gu2023_nips_waymax}      &{NeurIPS 2023}   & LT+DF &                             & 4.14$\pm$2.04 & 5.83$\pm$1.09 & \textcolor{gray}{79.58$\pm$24.98} \\
        PlanT\textsuperscript{$\dagger$}~\cite{renz2022plant}    & {CoRL 2023}   & LT    &                             & 2.29$\pm$0.11 & 3.08 $\pm$ 0.02 & 95.38$\pm$1.03 \\
        LatentDriver~\cite{xiao2024_arxiv_latentdriver}    & {ICRA 2025}  & LT    &                             & 2.33$\pm$0.13 & 3.17$\pm$0.04 & \textbf{99.57$\pm$0.1} \\
        \rowcolor{gray!15}
        ConsistencyPlanner (Ours)  &   & LT    &                             & \textbf{2.09} & \textbf{2.77} & 93.72 \\
       
        \bottomrule
    \end{tabular}
    }

    \vspace{2pt} 
  \begin{flushleft}
    \small 
    \item[] The dagger symbol ($\dagger$) on PlanT~\cite{renz2022plant} denotes results reproduced in our experiments using the officially released checkpoints.
  \end{flushleft}
\end{table*}

The comparative analysis with other methods on Waymax~\cite{Gu2023_nips_waymax} is presented in Table~\ref{tab:waymax_comparision}. Non-ego agents are controlled
by the Intelligent Driver Model (IDM)~\cite{treiber2000_idm}, as implemented
in previous works~\cite{Gu2023_nips_waymax,xiao2024_arxiv_easychauffeur}. The asterisk $*$ on PR indicates that this metric cannot be fairly compared under some methods due to the absence of drivable future under ‘Route’. This discrepancy is highlighted in gray. 
The baselines evaluated  include: (1)\textbf{Wayformer}~\cite{nayakanti2023_irca_wayformer}, a Transformer-based model for joint trajectory prediction of ego and surrounding agents, and \textbf{Wayformer-BC}, which uses the same encoder with an MLP decoder for ego action prediction; (2) \textbf{EasyChauffeur-PPO}~\cite{xiao2024_arxiv_easychauffeur}, which adopts the same encoder as ConsistencyPlanner, employs the reinforcement learning algorithm PPO to directly optimize control actions in a closed-loop setting, enabling policy learning through interaction with the environment. (3) \textbf{PlanT}~\cite{renz2022plant}, which adopts the same encoder as ConsistencyPlanner but replaces the decoder with an MLP without consistency modeling; and (4) \textbf{LatentDriver}~\cite{xiao2024_arxiv_latentdriver}, which models the latent future dynamics as a mixture distribution with sampling-based decoding. The mean and standard deviation are reported under three random seeds.

The experimental results demonstrate that the Waymax closed-loop simulation platform supports performance comparisons of advanced baseline algorithms in both continuous and discrete action spaces. However, since the platform does not provide open-source configuration scripts for discrete action settings, this section focuses on evaluating the closed-loop motion planning performance of the proposed ConsistencyPlanner model under a continuous action space. As shown in Table~\ref{tab:waymax_comparision}, the ConsistencyPlanner consistently outperforms all baseline methods in terms of OR and CR, despite achieving a slightly lower PR. Notably, it attains the lowest CR among all evaluated methods, and its OR is only marginally higher than that of the Wayformer-BC, which operates in a discrete action space.

\subsection{Ablation Study}
\label{sec:ablation}

This subsection further analyzes the potential impact of model architecture choices and generative model categories on closed-loop decision-making and planning performance through ablation experiments.

\subsubsection{Impact of Consistency Decoder Architecture}
The consistency decoder must effectively integrate multiple input features—including scene encodings, action features, consistency timesteps, and navigation paths—to generate precise and context-aware driving decisions. Our framework primarily employs Transformer blocks, utilizing self-attention and cross-attention mechanisms to achieve hierarchical feature fusion. For comparison, we also evaluate a simpler Multi-Layer Perceptron (MLP)-based architecture, where features are concatenated into a latent vector and processed through fully connected layers.

As shown in Table~\ref{tab:architecture_comparison}, the Transformer-based ConsistencyPlanner outperforms the MLP variant in closed-loop motion planning, achieving higher overlap rate (OR) and collision rate (CR) metrics, albeit with a marginally lower progress rate (PR). These results suggest that the Transformer’s attention mechanisms enhance feature fusion efficacy, particularly in improving safety-critical metrics.

\begin{table}[htbp]
  \centering
  \caption{Impact of consistency  decoder architecture on closed-loop motion planning performance}
  \begin{tabular}{c|c|cc|c}
    \toprule
    \textbf{Architecture} & \textbf{Action Space} & \textbf{OR$\downarrow$} & \textbf{CR$\downarrow$} &  \textbf{PR$\uparrow$} \\
    \midrule
    MLP & Waypoints & 2.39 & 2.82 & \textbf{94.99} \\
    Transformer Block & Waypoints & \textbf{2.09} & \textbf{2.77} & 93.72 \\
    \bottomrule
  \end{tabular}%
  \label{tab:architecture_comparison}%
\end{table}

\begin{table}[htbp]
  \centering
  \caption{Impact of generative model selection on closed-loop motion planning performance}
    \begin{tabular}{l|cc|c|c}
    \toprule
    \textbf{Method}  & \textbf{OR$\downarrow$} & \textbf{CR$\downarrow$} &  \textbf{PR$\uparrow$} & \small{Inference (ms) $\downarrow$} \\
    \midrule
    ConsistencyPlanner  & 2.09 & 2.77 & \textbf{93.72} & \textbf{15} \\
    DiffusionPlanner  & \textbf{2.07} & \textbf{2.76} & 93.63 & 122 \\
    \bottomrule
    \end{tabular}%
  \label{tab:generative_model_comparison}%
\end{table}%


\subsubsection{Impact of Generative Models}

ConsistencyPlanner employs fast-sampling consistency models as its core framework and conducts a systematic comparison with an enhanced diffusion model accelerated via the DPM-Solver algorithm. The performance comparison between these two generative models in closed-loop settings is summarized in Table~\ref{tab:generative_model_comparison}.

In terms of experimental setup, the diffusion model leverages the DPM-Solver with 10 sampling steps, significantly improving inference efficiency while preserving generation quality. In contrast, the consistency model enables rapid prediction through single-step inference. The results highlight the consistency model’s remarkable advantage in computational efficiency, reducing inference time to approximately 10\% of that required by the diffusion-based approach—clearly demonstrating the benefits of single-step sampling. With respect to safety metrics, the diffusion model exhibits slightly superior performance in indicators such as collision rate, largely attributable to its iterative refinement process, which facilitates a more thorough exploration of multimodal decision spaces. Nevertheless, the consistency model’s exceptional inference speed makes it more suitable for real-time autonomous driving scenarios. As a result, the consistency model achieves an effective trade-off between computational efficiency and safety, offering strong support for high-performance closed-loop planning.

\begin{figure*}[htp]
    \centering
    \includegraphics[width=1\textwidth]{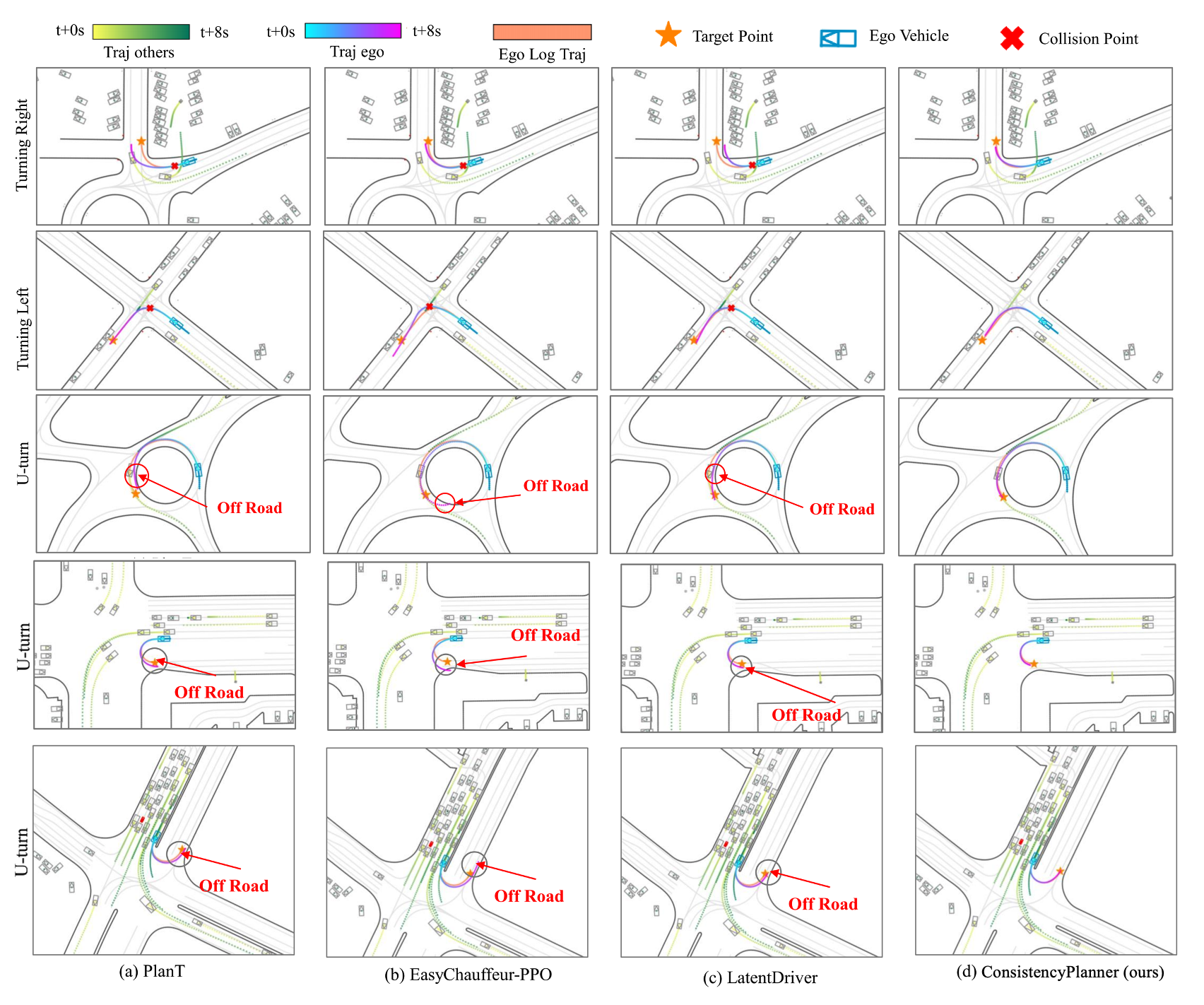}
    \caption{Visualization results of ConsistencyPlanner against other three methods in complex driving scenarios. }
    \label{cases}
\end{figure*}

\subsection{Visualization Results}

We evaluate the performance of four methods across distinct driving scenarios in Fig.~\ref{cases}. In the turning-right scenario, the other three methods collide with an oncoming left-turning vehicle due to their delayed response, whereas ConsistencyPlanner successfully negotiates the interaction through decisive maneuver planning. A similar pattern emerges in the turning-left scenario: while PlanT, EasyChauffeur, and Latent Driver fail to avoid contact with a straight-moving vehicle, ConsistencyPlanner maintains safe passage. The U-turn scenarios proves particularly challenging. Other methods exhibit temporary off-road deviations near the target point due to the aggressive roadway curvature, yet only ConsistencyPlanner successfully completes the maneuver while remaining within drivable boundaries. The quantitative results clearly demonstrate that our approach reliably maintains lane discipline and robustly prevents off-road excursions.

\section{Conclusion}
\label{sec:conclusion}

We present ConsistencyPlanner, a novel driving strategy designed for closed-loop planning in real-world autonomous vehicle scenarios. Our approach introduces a consistency model framework that effectively balances accurate multimodal behavior modeling with real-time decision-making capabilities. The proposed architecture integrates an attention mechanism during the ego vehicle's action decoding phase, enabling comprehensive fusion of conditional inputs with dynamic action information to produce precise planning decisions. This design results in future motion trajectories that are safe and comfortable, demonstrating the model’s strong potential for real-world deployment.

\section{Limitation}
\label{sec:limitation}

While the proposed ConsistencyPlanner achieves notable improvements in both off-road rate and collision rate metrics, its performance on the progress rate remains suboptimal—occasionally falling behind the baseline PlanT algorithm. Our in-depth qualitative analysis reveals that this limitation primarily arises from the model’s conservative motion tendencies, particularly at intersections, where it favors safety and collision avoidance over aggressive trajectory progression. To mitigate this, future work will explore the integration of classifier-free guidance, enabling the incorporation of progress-related objectives—such as the progress rate metric—directly into the sampling process through gradient-based optimization. This approach offers a principled way to balance safety with route completion, enabling finer control over the trade-offs between conservative and efficient planning behaviors.

\section{Acknowledgments}
This work is supported by the Beijing Natural Science Foundation under Grant 4242052 and the National Key Research and Development Program of China under Grant 2022YFA1004000.

{
\footnotesize
\bibliographystyle{ieeetr}
\bibliography{references.bib}
}
\vspace{12pt}

\end{document}